\pdfoutput=1
\documentclass[11pt]{article}

\usepackage[final]{acl}
\usepackage{placeins}
\usepackage{times}
\usepackage{latexsym}
\usepackage{float}
\usepackage{amsmath,amssymb}
\usepackage[T1]{fontenc}
\usepackage[utf8]{inputenc}

\usepackage{microtype}

\usepackage{inconsolata}

\usepackage{graphicx}

\usepackage{multirow}
\usepackage{arydshln} % 대시라인을 위한 패키지
\usepackage{booktabs} % 테이블에서 \toprule, \midrule, \bottomrule을 사용하기 위함
\usepackage{caption} % 캡션 사용을 위한 패키지 추가
\usepackage{natbib} % APA 스타일을 위해 natbib 패키지 사용
\usepackage{hyperref} % 링크를 위해 hyperref 패키지 사용
\usepackage{enumitem}
\usepackage{linguex}
\makeatletter

\title{Continuous Interpretive Steering for Scalar Diversity}
\author{Ye-eun Cho\\
  Sungkyunkwan University\\
  Seoul, South Korea\\
  \texttt{joyenn@skku.edu}}
\begin{document}
\maketitle
\begin{abstract}
Pragmatic inference is inherently graded. Different lexical items give rise to pragmatic enrichment to different degrees. Scalar implicature exemplifies this property through scalar diversity, where implicature strength varies across scalar items. However, evaluations of pragmatic inference in large language models (LLMs) often rely on prompt-based manipulations. Beyond prompt-level effects, this study introduces Continuous Interpretive Steering (CIS), a method that probes graded pragmatic interpretation by treating activation-level steering strength as a continuous experimental variable. To support this analysis, this study introduces a new dataset, GraSD, which encodes graded scalar diversity. Experiments on four LLMs show that uniform activation steering increases pragmatic interpretations globally but collapses item-level variation, whereas graded activation steering yields differentiated interpretive shifts aligned with scalar diversity grades. It indicates that graded sensitivity is encoded in the representation space and can be systematically recovered through controlled intervention. Together, CIS and GraSD provide a principled framework for evaluating graded pragmatic sensitivity in LLMs.\\
\end{abstract}

\section{Introduction}

\begin{figure}[t]
  \includegraphics[width=\columnwidth]{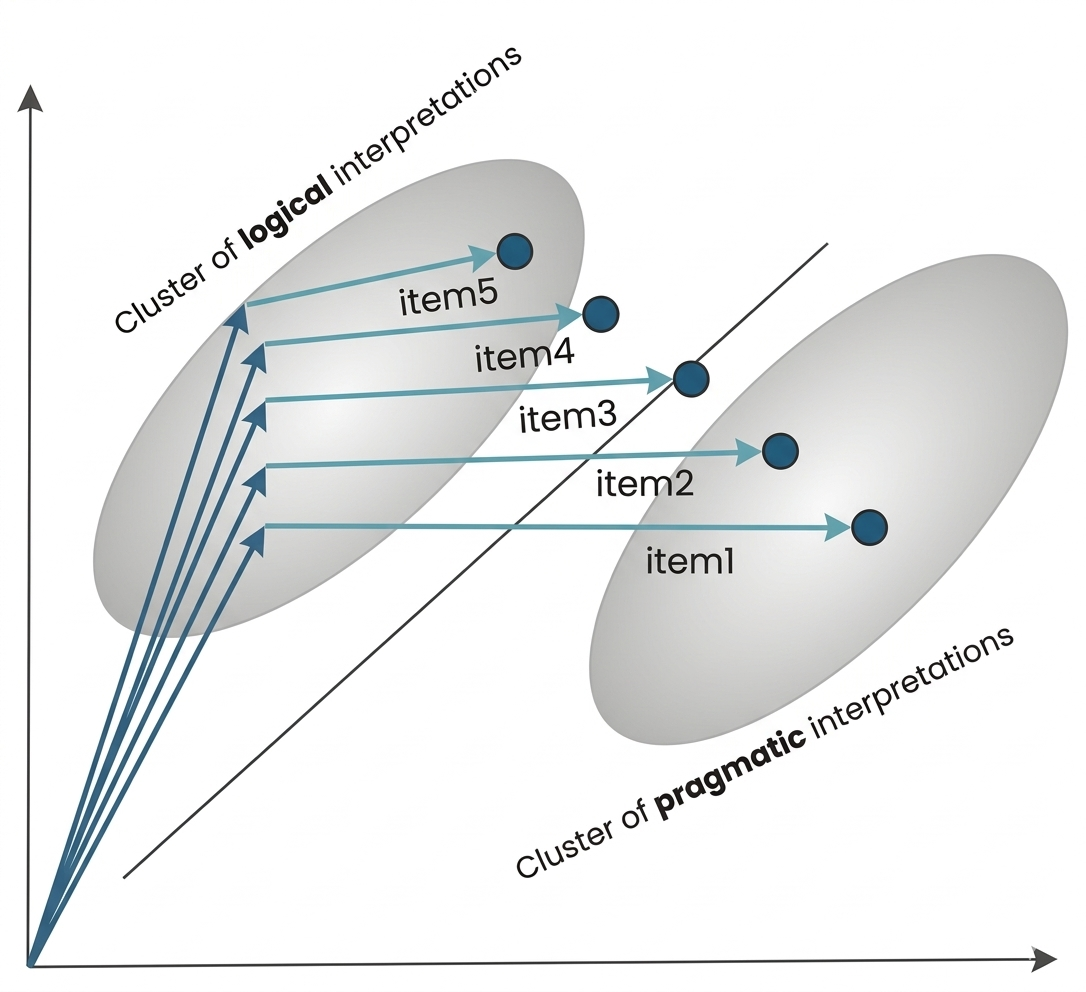}
  \caption{Continuous interpretive steering in representation space}
\end{figure}

\begin{figure}[t]
  \includegraphics[width=\columnwidth]{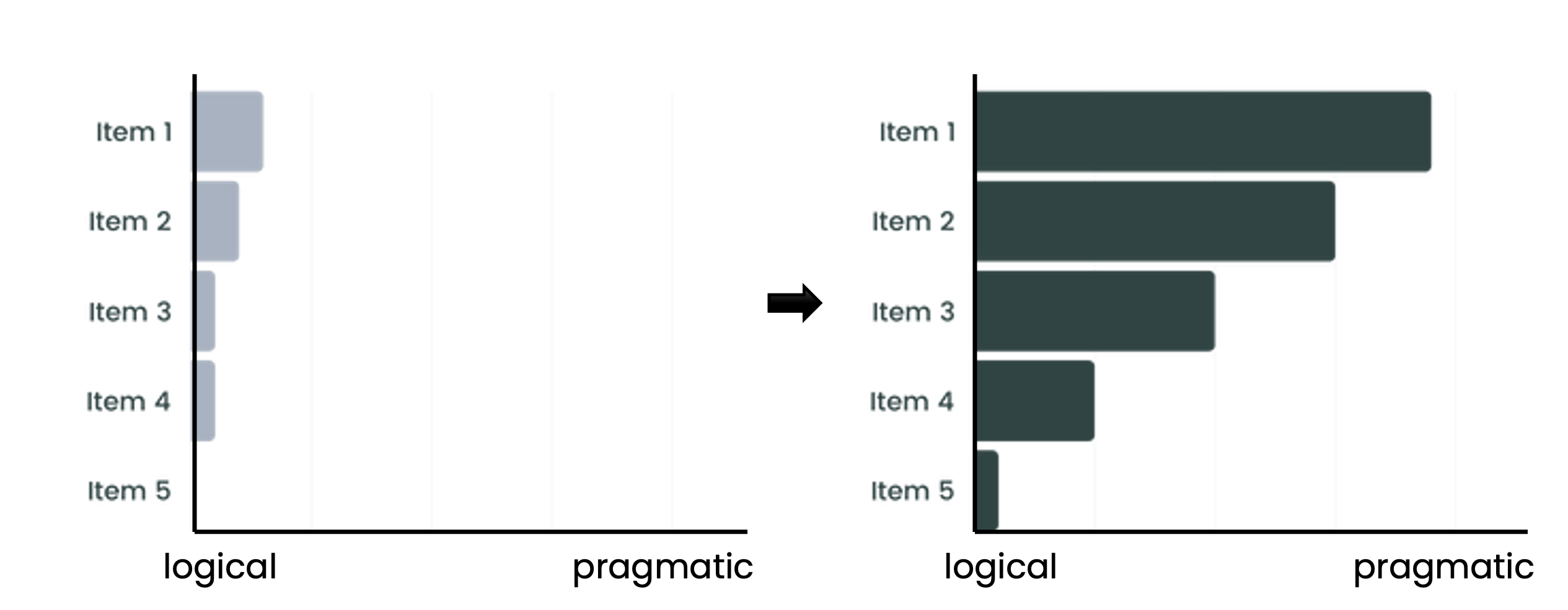}
  \caption{Graded changes in interpretive outcomes before and after steering}
\end{figure}

Pragmatic inference enables comprehenders to derive meanings that go beyond literal truth conditions. A canonical case is scalar implicature, where an utterance containing a weaker scalar term (e.g., \textit{some}) is often interpreted as negating stronger alternatives (e.g., \textit{not all}) when those alternatives are relevant \citep{horn1972semantic}. Scalar implicature has  therefore been widely used to evaluate whether large language models (LLMs) exhibit human-like pragmatic behavior, and recent work has reported both successes and failures depending on experimental framing and task format \citep{hu2023fine, cho2024pragmatic, wu2024rethinking, cho-maeng-2025-vision, tsvilodub2025non}.

However, most evaluations of pragmatic inference in LLMs rely primarily on prompt manipulations, varying instructions, exemplars, or question phrasing to elicit a pragmatic interpretation \citep{webson2022prompt, mielke2022reducing, turpin2023language, park2024pragmatic, cho2025prompting, cho-maeng-2025-vision}. While practical, prompt-based approaches are inherently ephemeral and can be brittle; measured pragmatic reasoning abilities may reflect sensitivity to surface form rather than stable internal representations. This concern is echoed by work arguing that pragmatic benchmarks can be highly prompt-sensitive, complicating mechanistic interpretation and cross-model comparison \citep{ruis2023goldilocks}.

Therefore, instead of inducing pragmatic behavior only through prompts, this work intervenes directly in the model’s internal activations at inference time. Inference-time steering methods modify hidden states during a forward pass without updating model weights, offering a lightweight alternative to fine-tuning and a more direct probe of model-internal computations \citep{turner2023steering, zou2023representation}. 

Importantly, pragmatic phenomena in human language are inherently graded. Specifically, in terms of scalar implicature, different scalar items trigger pragmatic enrichment at different degrees, which is known as scalar diversity \citep{van2016scalar}. Consequently, if LLMs are to approximate human-like pragmatic behavior, their pragmatic interpretations should likewise be modulated in an item-sensitive manner.

For this purpose, this study introduces Continuous Interpretive Steering (CIS) for scalar diversity. This approach constructs a steering direction in activation space and varies its strength continuously. Unlike prior steering work, which primarily relies on fixed intervention vectors to optimize downstream behaviors, the present approach treats steering strength as an experimental variable for analysis. Figure 1 illustrates the core idea of this approach in representation space and Figure 2 depicts the expected differences in interpretive outcomes before and after steering. To evaluate graded interpretive sensitivity across scalar items, a new dataset, GraSD (Graded Scalar Diversity), is constructed and used in this study.

\section{Background}
\subsection{Scalar Diversity}
Scalar implicature indicates that utterances containing weaker scalar terms (e.g., \textit{some}) are often interpreted as negating stronger alternatives (e.g., \textit{not all}) when the stronger alternatives are relevant \citep{horn1972semantic}. However, growing empirical evidence suggests that different scalar items give rise to pragmatic interpretations to different degrees (\citealp{van2016scalar, ronai2021pragmatic, ronai2024could, pankratz2021role}; see also \citealp{degen2015processing}). This phenomenon is known as scalar diversity.

For example, \citet{van2016scalar} showed that scalar items differ in how strongly they license implicatures: utterances with \textit{some} robustly invite \textit{not all}, whereas terms such as \textit{warm} rarely give rise to \textit{not hot}. Crucially, this variability persists even when contextual factors are carefully controlled. This graded nature of implicature strength across scalar items adopted from \citet{van2016scalar} is represented in Appendix~\ref{app:human}. Subsequent psycholinguistic studies have reinforced this tendency by showing that scalar implicature judgments often exhibit graded response patterns across various scalar items \citep{ronai2021pragmatic, ronai2024could, pankratz2021role}. These findings suggest that scalar diversity is a robust empirical property of human pragmatic behavior. Scalar implicature, in other words, is not simply present or absent, but instantiated to different degrees depending on the lexical items under consideration. These properties have direct implications for modeling pragmatic inference.

Recent work attempting to elicit human-like pragmatic behavior often aim to shift models away from literal interpretations toward pragmatic ones \citep{hu2023fine, cho2024pragmatic, wu2024rethinking, cho-maeng-2025-vision, tsvilodub2025non}. However, such approaches risk collapsing scalar diversity by treating pragmatic enrichment as a uniform target. If a model is pushed indiscriminately toward pragmatic interpretation, it may overgenerate enriched readings in cases where humans exhibit only weak or marginal implicatures.

From the perspective of scalar diversity, human-like pragmatic reasoning requires the capacity to modulate the strength of pragmatic enrichment at the level of individual items, reflecting the graded patterns observed in human judgments. This motivates an evaluation framework that goes beyond binary success or failure and instead examines whether models can reproduce item-level variation in pragmatic interpretation.

\subsection{Internal Activation Interventions}
While LLMs effectively elicit certain behaviors with input-level manipulations, such as prompt engineering or task rephrasing, these approaches operate indirectly on the model’s internal computations and are often sensitive to surface form. An alternative approach intervenes directly on a model’s internal activations at inference time \citep{turner2023steering, zou2023representation, wang2024semantics}. Inference-time steering methods modify hidden states during the forward pass of a pretrained model, without updating its parameters. This paradigm has been explored under various labels, including activation engineering and representation engineering, and has been shown to enable lightweight control over model outputs with minimal computational cost \citep{turner2023steering, zou2023representation}.

Most existing steering approaches construct a fixed intervention vector, often derived from contrastive examples, and apply it uniformly to bias model behavior toward a desired outcome \citep{turner2023steering, wang2024semantics, bayat2025steering, hojer2025improving, neplenbroek-etal-2025-reading, soo2025interpretable, suri2025mitigating}. These methods have primarily been evaluated on downstream objectives such as sentiment control, truthfulness, or alignment, where the goal is to optimize performance rather than to analyze internal structure. Consequently, steering is typically treated as a one-shot control mechanism rather than as an experimental variable.

Recent work has begun to incorporate gradient-based signals, such as attribution scores, to construct more targeted steering directions \citep{nguyen2025grains}. However, even in these settings, steering strength is rarely varied systematically, and evaluation focuses on end-task improvements rather than on graded shifts in interpretation.

In contrast, the present study treats steering as a probing tool. By varying steering strength continuously, internal intervention is used to trace how interpretations change across scalar items. This perspective aligns naturally with scalar diversity. If pragmatic inference is internally represented as a graded state, then small changes in internal activations should produce smooth, item-sensitive shifts. Steering thus provides a means of probing the geometry of pragmatic inference within LLMs, complementing traditional prompt-based evaluations.

\section{Dataset}
To evaluate graded pragmatic behavior in LLMs, this study constructs a dataset, called GraSD. GraSD consolidates <weak, strong> item pairs from four prior studies that investigate scalar diversity under controlled experimental settings as follows:

\vspace{1em}
\begin{itemize}[noitemsep, topsep=0pt]
  \item 43 pairs \citep{van2016scalar}
  \item 43 pairs \citep{ronai2021pragmatic}
  \item 50 pairs \citep{pankratz2021role}
  \item 60 pairs \citep{ronai2024could}
\end{itemize}
\vspace{1em}

After removing duplicate entries across sources, a total of 121 <weak, strong> pairs were retained. These pairs serve as the basis for constructing experimental sentences consisting of anchor, logical, and pragmatic variants as in Table 1. For each scalar pair, the anchor sentence contains the weak scalar term and represents the baseline utterance from which interpretations are derived. The logical variant replaces the weak term with its stronger alternative, thereby expressing a semantically stronger proposition within the same scale. The pragmatic variant encodes the scalar implicature associated with the anchor, typically realized as the negation of the stronger alternative.\footnote{In formal semantics, weak terms (e.g., \textit{some}) are compatible with stronger cases (e.g., \textit{all}), as they express existential quantification \citep{horn1972semantic}. The logical variant therefore represents a stronger proposition within the scale, while the pragmatic variant corresponds to the negation of this stronger alternative.}

\begin{table}[t]
\centering
\small
\begin{tabular}{ll}
\toprule
Type & Sentence \\
\midrule
Anchor & Some pets prefer to sleep in the sun. \\
Logical & All pets prefer to sleep in the sun. \\
Pragmatic & Not all pets prefer to sleep in the sun. \\
\bottomrule
\end{tabular}
\caption{Example of experimental sentences for the pair <some, all>.}
\end{table}

To enable large-scale evaluation, the 121 base item pairs were augmented into a range of contextualized sentence instances. A theory-driven, constraint-based augmentation strategy was employed, providing the GPT-4o model \citep{openai2024gpt4o} with linguistically motivated constraints derived from scalar implicature theory, which is inspired by \citet{su2025cas}. Based on these theoretical constraints, the model generated diverse contextual instances while consistently preserving the intended forms among the anchor, logical, and pragmatic sentences. Compared to rule-based methods, this approach produces richer and more varied contexts; and compared to unconstrained generation, it offers greater control over the pragmatic conditions under which scalar implicatures may arise.
Through this approach, by producing 100 instances per scalar item pair, a total of 121,000 sentence instances were created. The scalar item pairs included in GraSD and  the prompt used for the data augmentation are listed in Appendix~\ref{app:data}. The complete dataset, experimental code and results are publicly available.\footnote{\url{https://github.com/joyennn/CIS}}

\section{Methods}
\subsection{Models}
Experiments are conducted on four open-weight transformer-based language models: LLaMA3 \citep{grattafiori2024llama}, Qwen2 \citep{team2024qwen2}, Gemma2 \citep{team2024gemma}, and OLMo \citep{groeneveld2024olmo}. These models were selected to cover a range of architectures and training regimes while maintaining full access to internal representations required for activation-level steering.

\subsection{Activation-level Steering}
This study implements activation-level steering as an intervention defined in the model’s activation space at inference time, without updating model parameters. Rather than modifying hidden states during the forward pass, the approach operates on extracted internal representations, enabling controlled manipulation of activation geometry while preserving the model’s original computation.

Let $h_l(x)\in\mathbb{R}^d$ denote the hidden activation produced by layer $l$ for an input sentence $x$. For each input, hidden representations are extracted from a fixed set of $k$ transformer layers and concatenated to form a multi-layer representation vector $h(x)\in\mathbb{R}^{kd}$. Steering is defined by an intervention vector $v\in\mathbb{R}^{kd}$, which represents a direction in this aggregated activation space. Given a scalar coefficient $\alpha\in\mathbb{R}$, the steered representation is defined as:

\begin{equation}
h'(x) = h(x) + \alpha v
\label{eq:steering}
\end{equation}

Here, $v$ specifies the direction of representational change, while $\alpha$ controls the magnitude of the intervention. The intervention vector is computed once and shared across all inputs, ensuring that variation in model behavior arises from differences in steering strength rather than from item-specific directions.

This approach is closely related to prior work on activation and representation engineering \citep{turner2023steering, wang2024semantics, bayat2025steering, hojer2025improving, neplenbroek-etal-2025-reading, soo2025interpretable, suri2025mitigating}, but differs in its use of steering as an analytical probe rather than as a control mechanism for downstream task optimization.

\subsection{Continuous Interpretive Steering}
Building on the activation-level steering framework described above, Continuous Interpretive Steering (CIS) is introduced as a method that treats steering strength as a continuous experimental variable for probing graded pragmatic interpretation. Rather than using steering to enforce a fixed behavioral outcome, controlled variation in steering magnitude is exploited to examine how internal representations shift across interpretive alternatives.

Each dataset item consists of an anchor sentence and its corresponding logical and pragmatic variants. For a given item, multi-layer representations are extracted for all three sentence types under fixed model parameters. Using these representations, a pragmatic direction in activation space is defined as the difference between the pragmatic and logical representations.

The anchor representation is then steered along this pragmatic direction by varying the steering coefficient $\alpha$ over a continuous range. This yields a family of steered anchor representations corresponding to different degrees of internal displacement toward the pragmatic interpretation. Crucially, the intervention direction is held constant across items, and only the magnitude of steering is varied, allowing item-level sensitivity to be assessed independently of direction-specific effects.

In practice, the pragmatic direction is estimated from a limited subset of instances per item (three instances per item pair), indicating that a stable interpretive direction can be extracted from a relatively small sample of data. This direction is subsequently held constant across all items during analysis.

\section{Experiments}
\subsection{Uniform Activation Steering}
Uniform activation steering applies a single steering direction with a fixed magnitude to all items, regardless of their lexical properties or interpretive characteristics. This condition serves as a control condition that probes the effect of applying activation steering without differentiation across items.

Concretely, a steering vector is added to the model’s internal representations at a fixed layer, scaled by a constant coefficient $\alpha$. The same steering coefficient is used for every item in the dataset, ensuring that all representations are shifted by an equal amount along the same direction in representation space.

The purpose of this setup is to examine the effects of a uniform representational shift on interpretive outcomes, while preserving a simple and controlled intervention scheme. By holding steering strength constant, this condition provides a reference point for assessing whether activation steering alone can induce systematic changes in interpretation.

\subsection{Graded Activation Steering}
Graded activation steering extends the uniform steering setup by allowing steering strength to vary across items, while keeping the steering direction identical to that used in uniform activation steering. This condition introduces controlled variation in steering magnitude as an experimental variable.

Specifically, items are assigned to discrete grade conditions (A–E), each of which is associated with a different steering coefficient $\alpha$. For example, grade A corresponds to the strongest steering magnitude and grade E to the weakest. Importantly, grades are used solely as a mechanism for modulating steering strength and do not encode semantic or interpretive categories in themselves. Items in different grades therefore experience different magnitudes of representational shift along the same steering direction.

\subsection{Analysis Strategies}

Interpretive preference is operationalized in representational terms. For each steered anchor representation, cosine similarity to the pragmatic and logical representations of the same item is computed. An item is considered to favor the pragmatic interpretation when the steered anchor becomes more similar to the pragmatic variant than to the logical variant. By tracking how this preference changes as a function of $\alpha$, a graded profile of interpretive sensitivity is obtained for each item.

Differences between steering conditions are evaluated using two complementary statistical tests. First, Wilcoxon signed-rank tests \citep{wilcoxon1945individual} are used to assess whether activation steering induces a significant shift in interpretive preference relative to the baseline condition. Second, Spearman rank correlations \citep{spearman1987proof} are employed to examine whether item-level interpretive preferences observed in the baseline condition are preserved under steering, thereby assessing the degree to which steering maintains graded sensitivity across items.

\section{Results}

\begin{figure}[t]
  \includegraphics[width=\columnwidth]{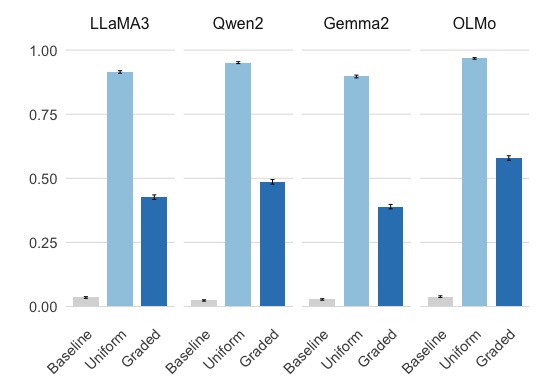}
  \caption{Proportion of pragmatic interpretations for baseline, uniform activation steering, and graded activation steering conditions across LLMs}
\end{figure}

\subsection{Overview of Interpretive Patterns}
Across all four models, clear differences emerge in interpretive behavior across steering conditions as shown in Figure 3, which summarizes the proportion of instances favoring the pragmatic interpretation across baseline, uniform and graded activation steering without distinguishing individual items. 

Specifically, across the four models, pragmatic interpretations are rare in the baseline condition, indicating that, in the absence of steering, model preferences are not biased toward pragmatic reasoning. Introducing activation steering leads to a substantial shift in overall interpretive behavior. Under uniform activation steering, the proportion of pragmatic interpretations increases markedly relative to baseline. Under graded activation steering, pragmatic interpretations remain substantially more frequent than in the baseline condition, although the overall proportion is reduced relative to uniform steering. 

The increase from baseline to uniform conditions suggests that activation steering exerts a robust global effect on interpretive preferences across models. The lower proportion under graded activation steering compared to uniform activation steering implies that modulation of steering strength redistributes interpretive sensitivity. These patterns are more strongly captured in item-level proportions of pragmatic interpretations under each steering condition, which are reported in Appendix~\ref{app:item_level}.

\subsection{Effects of Uniform Activation Steering}

\begin{figure}[t]
  \includegraphics[width=\columnwidth]{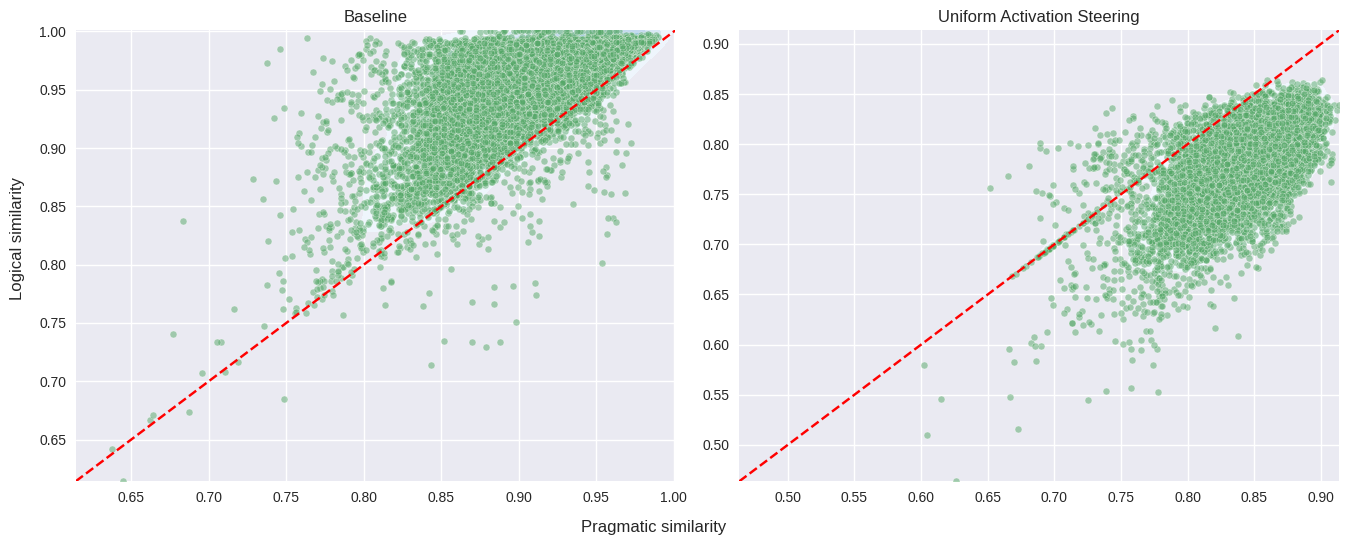}
  \caption{Scatter plots illustrating the relationship between pragmatic similarity (x-axis) and logical similarity (y-axis) under the baseline condition (left) and uniform activation steering (right) for OLMo}
\end{figure}

\begin{figure*}[t]
  \centering
  \includegraphics[width=\textwidth]{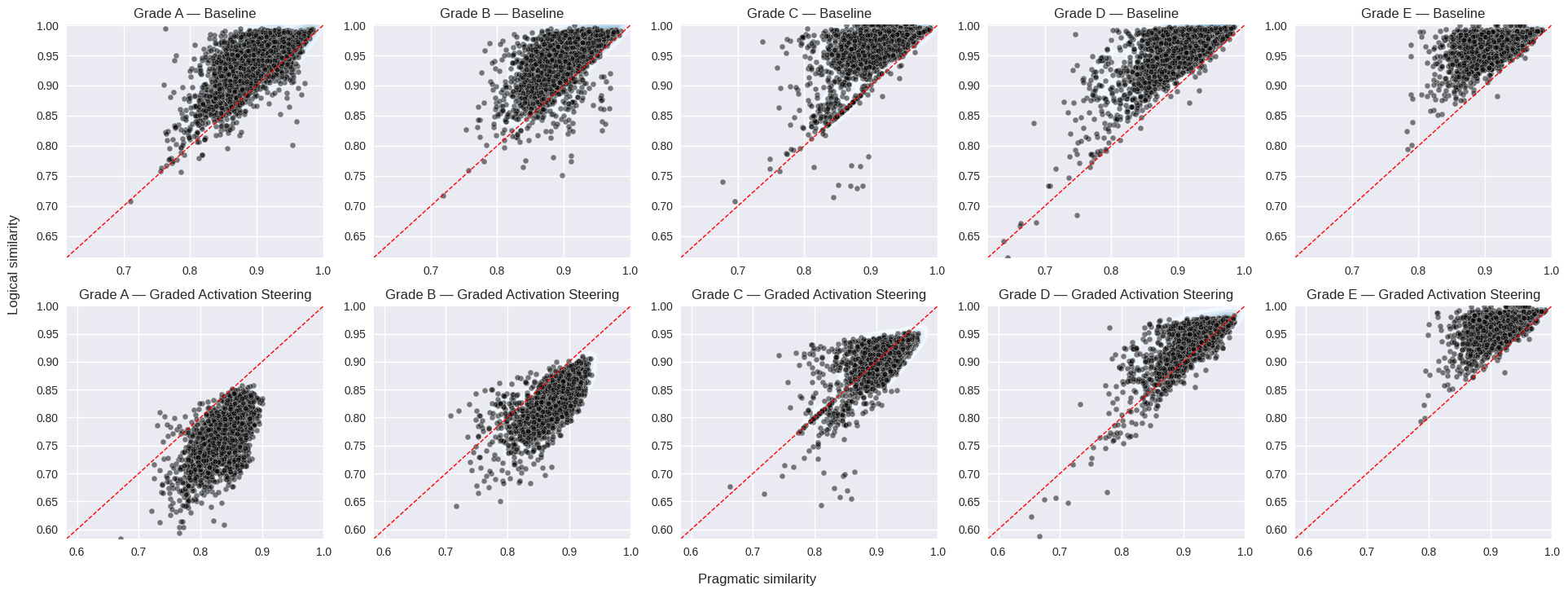}
  \caption{Scatter plots illustrating the relationship between pragmatic similarity (x-axis) and logical similarity (y-axis) under the baseline condition (left) and uniform activation steering (right) for OLMo}
  \label{fig:scatter_olmo}
\end{figure*}

Uniform activation steering, implemented with a fixed steering magnitude, induces systematic changes in interpretive preference across models. Results from the Wilcoxon signed-rank tests in Table 2 show that uniform activation steering leads to a robust and statistically significant shift in interpretive behavior for all four models (\textit{p}s < .001). This indicates that uniform activation steering consistently increases the overall proportion of pragmatic interpretations relative to the unsteered baseline.

However, this global shift is not accompanied by sensitivity to item-level variation. As shown by the Spearman rank correlations in Table 3, correlations between baseline interpretive preferences and those observed under uniform steering are weak and non-significant across all models (\textit{p}s = .78, .91, .60, .81). This suggests that uniform activation steering does not preserve the relative ordering of items, but increases pragmatic responses in a relatively homogeneous manner across items.

Figure 4 illustrates this pattern for a representative model, showing a uniform shift in pragmatic similarity under uniform activation steering relative to the baseline. Corresponding plots for the remaining models are provided in Appendix~\ref{app:uniform}.

These results characterize uniform activation steering as a coarse-grained intervention. While it reliably induces a global shift toward pragmatic interpretations, it provides limited insight into graded interpretive sensitivity or differential item effects. This limitation motivates the use of graded activation steering.

\begin{table}[t]
  \centering
  \resizebox{\columnwidth}{!}{%
  \begin{tabular}{lcccc}
    \toprule
     & \multicolumn{2}{c}{\textbf{Uniform}} & \multicolumn{2}{c}{\textbf{Graded}} \\
    \textbf{Model} & \textbf{$W$} & \textit{p}-value & \textbf{$W$} & \textit{p}-value \\
    \midrule
    LLaMA3 & 5.0 & \textbf{$<.001$} & 29.0 & \textbf{$<.001$} \\
    Qwen2  & 1.0 & \textbf{$<.001$} & 27.0 & \textbf{$<.001$} \\
    Gemma2 & 6.0 & \textbf{$<.001$} & 28.5 & \textbf{$<.001$} \\
    OLMo   & 2.0 & \textbf{$<.001$} & 32.0 & \textbf{$<.001$} \\
    \bottomrule
  \end{tabular}}
  \captionof{table}{Wilcoxon signed-rank test results comparing uniform and graded activation steering across LLMs}
  \label{tab:wilcoxon}
\end{table}

\begin{table}[t]
  \centering
  \resizebox{\columnwidth}{!}{%
  \begin{tabular}{lcccc}
    \toprule
     & \multicolumn{2}{c}{\textbf{Uniform}} & \multicolumn{2}{c}{\textbf{Graded}} \\
    \textbf{Model} & \textbf{$\rho$} & \textit{p}-value & \textbf{$\rho$} & \textit{p}-value \\
    \midrule
    LLaMA3 & 0.02  & 0.78           & 0.38  & \textbf{$<.001$} \\
    Qwen2  & -0.01 & 0.91           & 0.27  & \textbf{$<.01$}  \\
    Gemma2 & 0.05  & 0.60           & 0.39  & \textbf{$<.001$} \\
    OLMo   & -0.02 & 0.81           & 0.38  & \textbf{$<.001$} \\
    \bottomrule
  \end{tabular}}
  \captionof{table}{Spearman rank correlations comparing uniform and graded activation steering across LLMs}
  \label{tab:spearman}
\end{table}

\subsection{Effects of Graded Activation Steering}
Unlike uniform activation steering, graded activation steering modulates steering strength continuously across items. Results from the Wilcoxon signed-rank tests in Table 2 show that graded activation steering also produces a statistically significant shift in interpretive behavior relative to the baseline condition across all four models (\textit{p}s < .001). Compared to uniform steering, graded steering produces larger Wilcoxon statistics (\textit{W}), reflecting more differentiated item-level responses rather than uniformly strong shifts across items.

This indication is reinforced by Spearman rank correlations in Table 3. Correlations between baseline interpretive preferences and those observed under graded steering are moderate and statistically significant for all models (\textit{p}s < .01, .001). This suggests that graded activation steering preserves item-level sensitivity in interpretive preferences.

Figure 5 illustrates this pattern for a representative model, showing that graded steering yields differentiated responses across graded items, reflecting graded interpretive sensitivity. Corresponding plots for the remaining models are provided in Appendix~\ref{app:graded}.

These results characterize graded activation steering as a fine-grained intervention. While it induces a reliable global shift toward pragmatic interpretations, it does so in a manner that preserves item-level structure and relative interpretive differences. This property makes graded activation steering particularly well-suited for probing pragmatic interpretive behavior.

\subsection{Direct Comparison: Uniform vs Graded Steering}
A direct contrast between uniform and graded activation steering reveals qualitatively distinct interpretive patterns. Although both steering regimes produce statistically significant deviations from the baseline condition, their effects differ fundamentally in terms of distributional structure and item-level organization.

At the aggregate level, both uniform and graded steering yield robust Wilcoxon signed-rank test results, indicating reliable shifts toward pragmatic interpretations relative to baseline. However, as shown in Sections 6.2 and 6.3, statistical significance alone does not distinguish between homogeneous perturbation and structurally organized change. Additional descriptive statistics for grade-wise deviations under uniform and graded steering are reported in Appendix~\ref{app:deviation}.

A direct comparison therefore requires examining how steering effects are distributed across individual items. Figure 6 visualizes the distribution of item-level changes in pragmatic interpretation rate ($\Delta$ relative to baseline) under the two steering regimes. Under uniform activation steering, the distribution is sharply skewed toward large positive values, with most items exhibiting similarly strong increases in pragmatic interpretation. This pattern indicates a broad and relatively homogeneous displacement of model behavior, consistent with a coarse-grained intervention that affects items uniformly regardless of their baseline interpretive tendencies. Corresponding histograms for the remaining models are provided in Appendix~\ref{app:change}.

In contrast, graded activation steering produces a substantially broader distribution of item-level changes. While many items still show increased pragmatic interpretation, the magnitude of change varies widely, ranging from negligible shifts to large positive deviations. This heterogeneous distribution reflects differentiated item-level responses. 

Importantly, this variability is not random. As indicated by the significant Spearman rank correlations reported in Table 3, the magnitude of item-level responses under graded steering increases systematically along the pre-defined A–E grade hierarchy, with higher-grade items exhibiting larger deviations from baseline than lower-grade items. 

In addition, these grade-wise results confirm that graded activation steering induces interpretive change in a manner that respects and exploits the underlying A–E grade structure. By aligning steering effects with a theoretically motivated hierarchy of pragmatic strength, graded steering enables the emergence of fine-grained interpretive organization that cannot be achieved through fixed-magnitude interventions.

\section{Conclusion}

\begin{figure}[t]
  \includegraphics[width=\columnwidth]{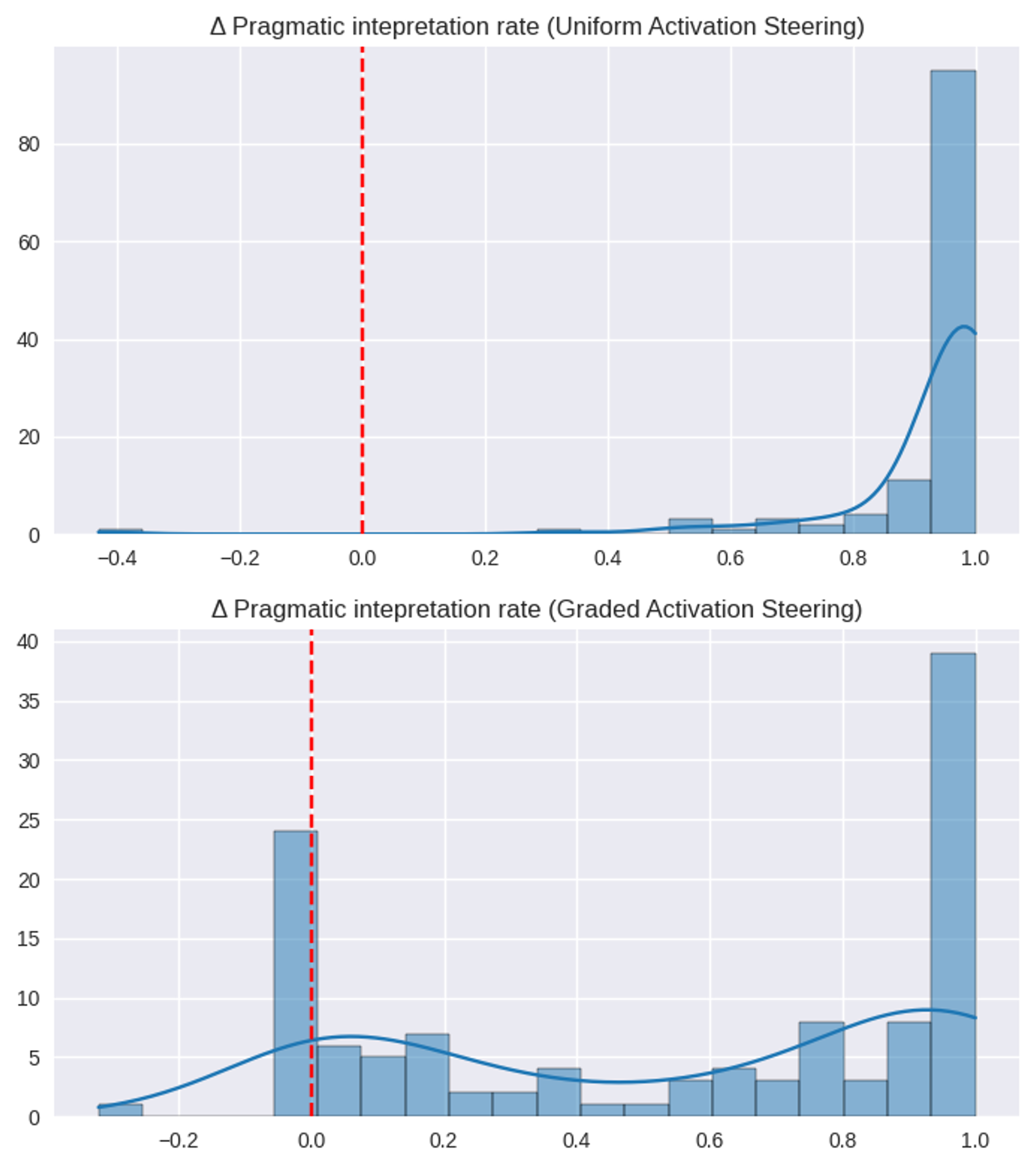}
  \caption{Histograms showing item-level changes in pragmatic interpretation rate ($\Delta$ relative to baseline) under uniform (top) and graded (bottom) activation steering for OLMo. }
\end{figure}

This study proposed Continuous Interpretive Steering (CIS) as a method for probing graded pragmatic interpretation in LLMs, with a particular focus on scalar diversity. Specifically, CIS conceptualizes pragmatic interpretation as a graded internal phenomenon that can be examined by systematically varying steering strength in activation space.

Across four LLMs, a clear contrast was observed between uniform and graded activation steering. Uniform steering reliably increased the overall rate of pragmatic interpretations, but did so in a largely homogeneous manner that obscured item-level differences. In contrast, graded activation steering yielded differentiated interpretive shifts that aligned with independently established scalar diversity grades and preserved the sensitivity to item-specific variation.

These findings highlight a limitation of aggregate evaluations of pragmatic behavior. Global increases in pragmatic interpretations alone do not capture structured pragmatic competence. Instead, preserving graded, item-level variation provides a more informative signal of how pragmatic distinctions are organized internally. By treating steering strength as a continuous experimental variable, CIS enables a fine-grained analysis of how internal representations shift between logical and pragmatic interpretations. It also indicates that the representational space encodes graded sensitivity that can be systematically recovered through controlled intervention.

In support of this analysis, the study introduced GraSD, a new dataset that encodes graded pragmatic strength across diverse scalar item pairs. By pairing GraSD with graded activation steering, this work provides both a methodological framework and a publicly available resource for evaluating graded pragmatic sensitivity in LLMs.

While the present study focused on scalar implicature, the proposed approach is not limited to this domain. Many pragmatic phenomena exhibit graded interpretive profiles across lexical items and contexts. CIS, therefore, offers a general tool for investigating the internal organization of pragmatic inference in LLMs.

\section*{Limitations}
This study has several limitations. First, the analysis focuses exclusively on scalar implicature, and the extent to which the proposed approach generalizes to other pragmatic phenomena remains an open question. Second, the construction of graded activation steering relies on a fixed steering direction estimated from a limited subset of instances, which may not capture the full variability of pragmatic representations across contexts. Finally, interpretive preference is operationalized through representational similarity rather than explicit behavioral judgments, which provides a controlled internal measure but may not fully reflect end-user interpretations. Future work may address these limitations by extending the approach to a broader range of pragmatic phenomena, model architectures, and evaluation measures.
\\
\nocite{*}
\bibliography{custom}

\appendix
\clearpage
\FloatBarrier
\section{Scalar Diversity in Human Judgments}
\label{app:human}
\begin{figure}[H]
  \includegraphics[width=\columnwidth]{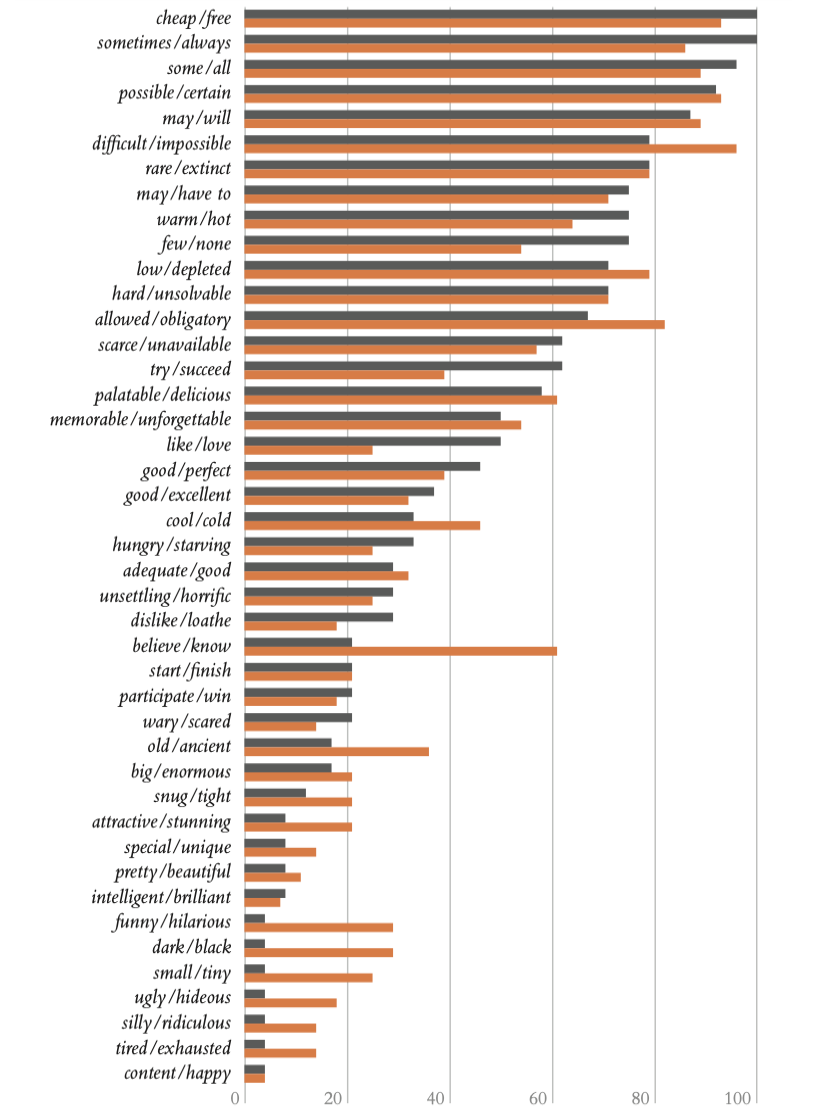}
  \caption{Graded scalar diversity observed in human judgments, adapted from van Tiel et al. (2016). Different scalar items exhibit systematically varying strengths of pragmatic interpretation}
\end{figure}

\begin{figure}[t]
\section{Scatter Plots for Uniform Activation Steering}
\label{app:uniform}
  \includegraphics[width=\columnwidth]{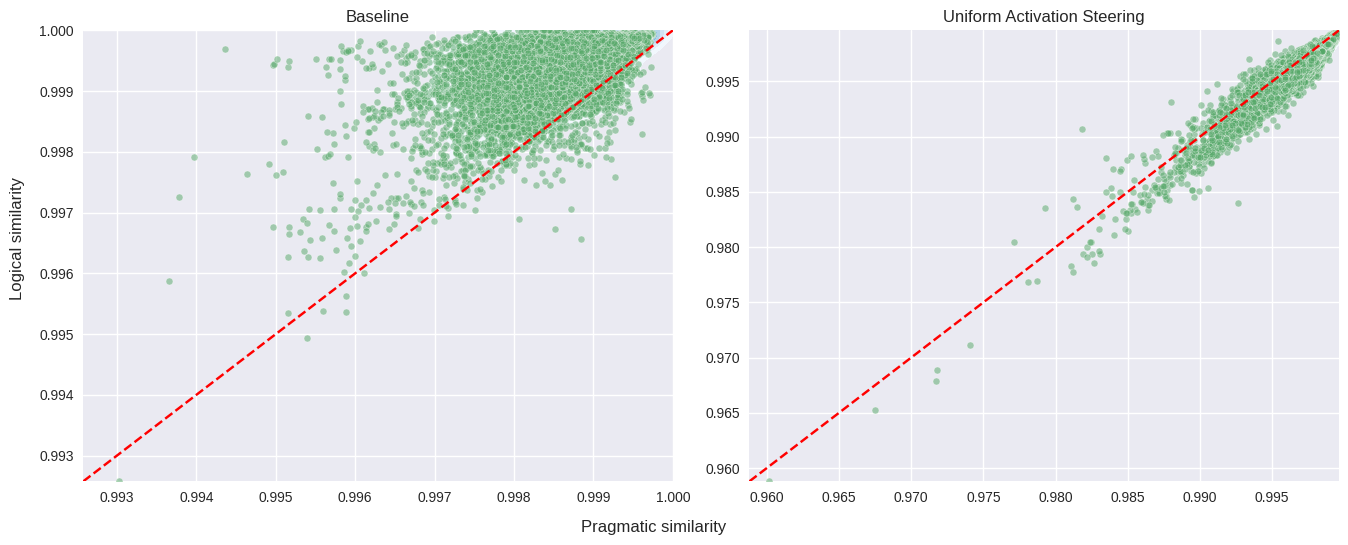}
  \caption{Scatter plots illustrating the relationship between pragmatic similarity (x-axis) and logical similarity (y-axis) under the baseline condition (left) and uniform activation steering (right) for LLaMA3}
\end{figure}

\begin{figure}
  \includegraphics[width=\columnwidth]{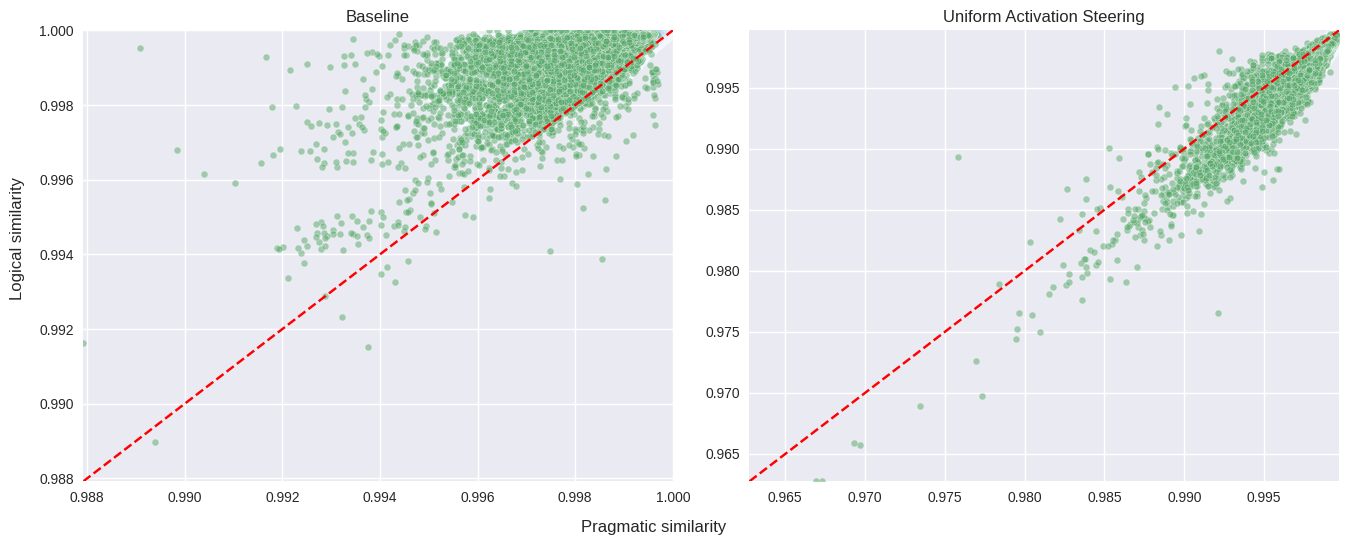}
  \caption{Scatter plots illustrating the relationship between pragmatic similarity (x-axis) and logical similarity (y-axis) under the baseline condition (left) and uniform activation steering (right) for Qwen2}
\end{figure}

\begin{figure}
  \includegraphics[width=\columnwidth]{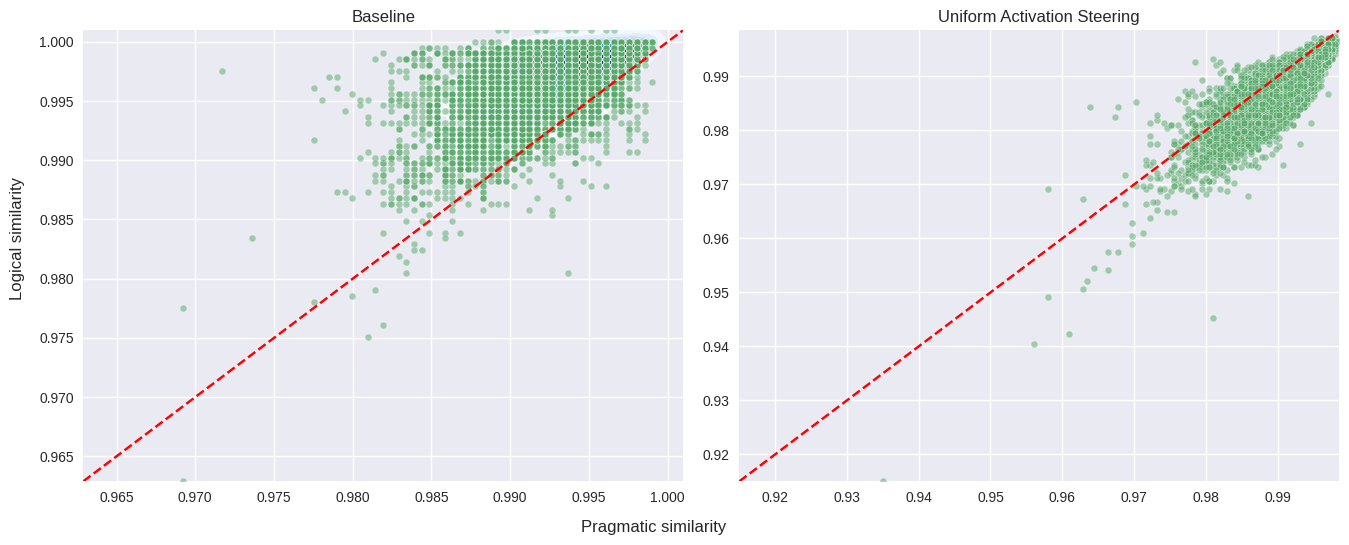}
  \caption{Scatter plots illustrating the relationship between pragmatic similarity (x-axis) and logical similarity (y-axis) under the baseline condition (left) and uniform activation steering (right) for Gemma2}
\end{figure}

\begin{figure}
  \includegraphics[width=\columnwidth]{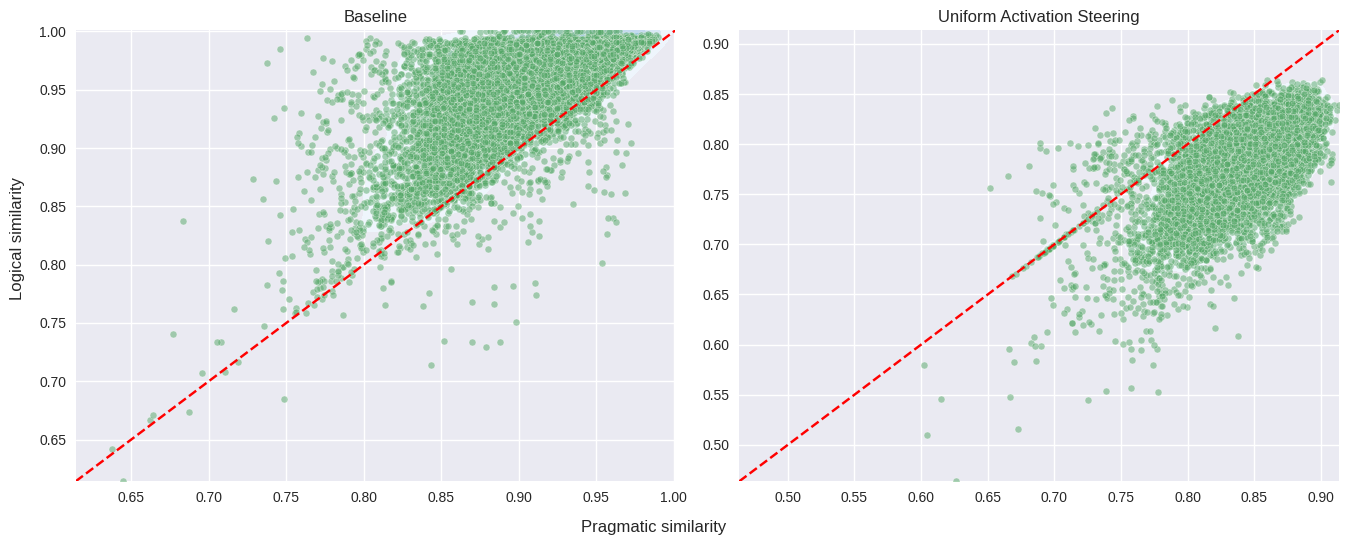}
  \caption{Scatter plots illustrating the relationship between pragmatic similarity (x-axis) and logical similarity (y-axis) under the baseline condition (left) and uniform activation steering (right) for OLMo}
\end{figure}

\begin{figure*}
\section{Items in GraSD dataset}
\label{app:data}
\subsection{Scalar Item Pairs from Source Studies}
  \includegraphics[width=\textwidth]{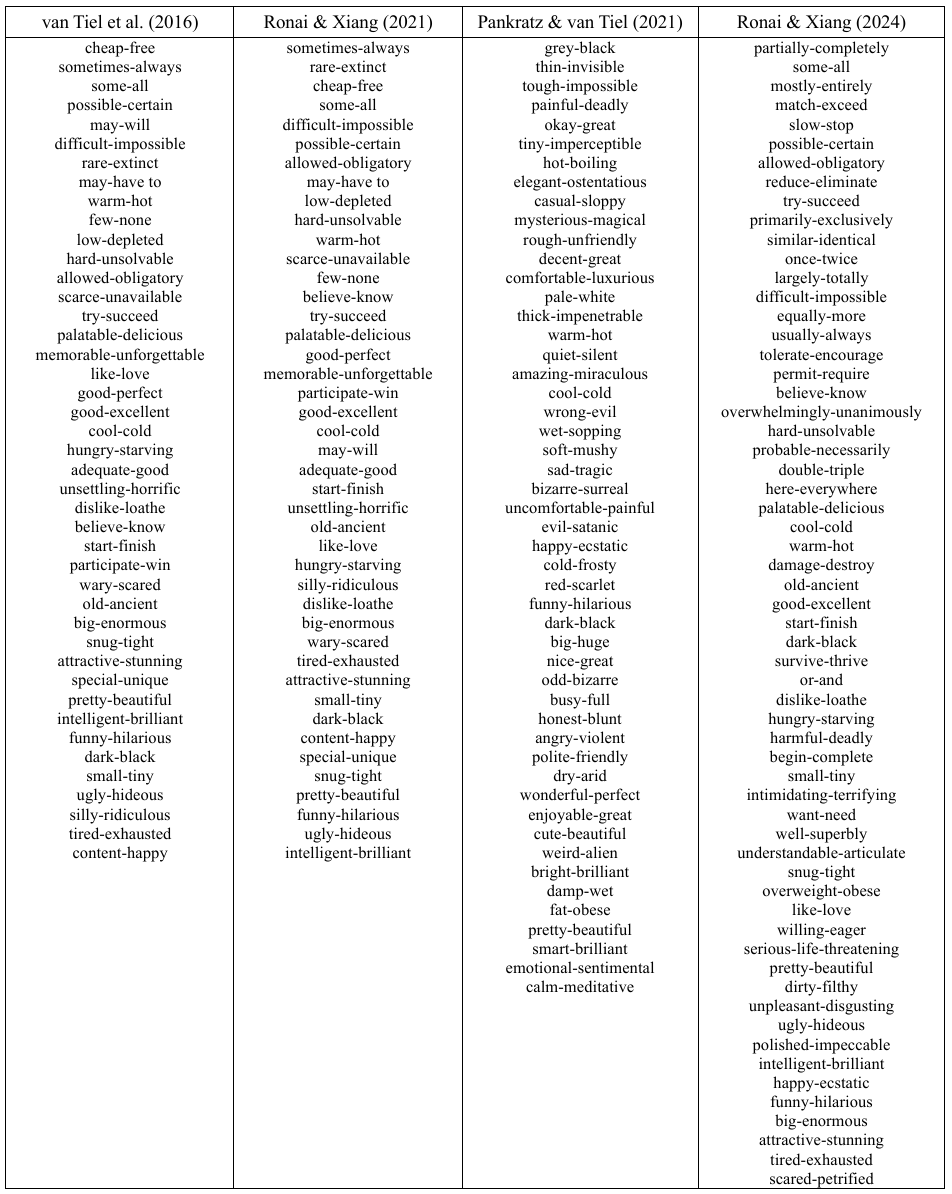}
\end{figure*}

\begin{figure*}
\subsection{Graded Scalar Item Pairs (A-E) in GraSD dataset}
  \includegraphics[width=\textwidth]{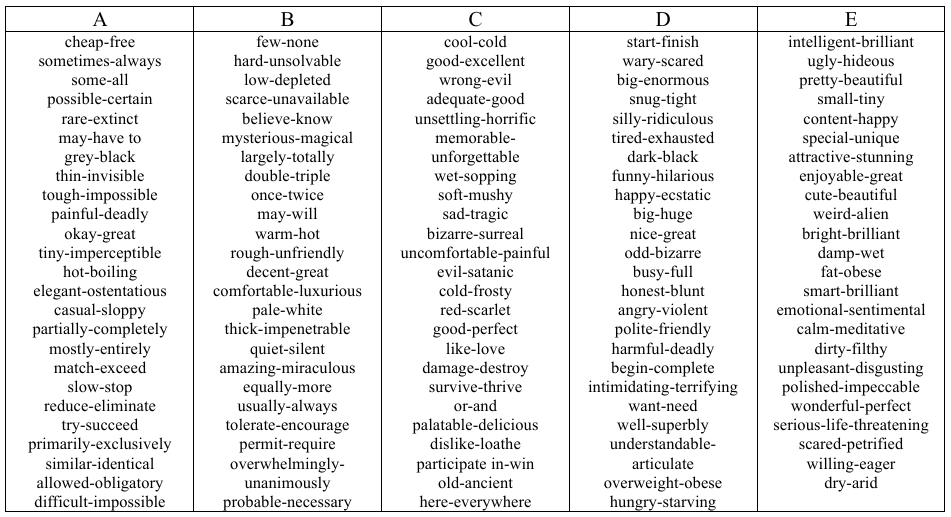}
\end{figure*}

\begin{figure*}
\subsection{Prompt for Data Augmentation}
  \includegraphics[width=\textwidth]{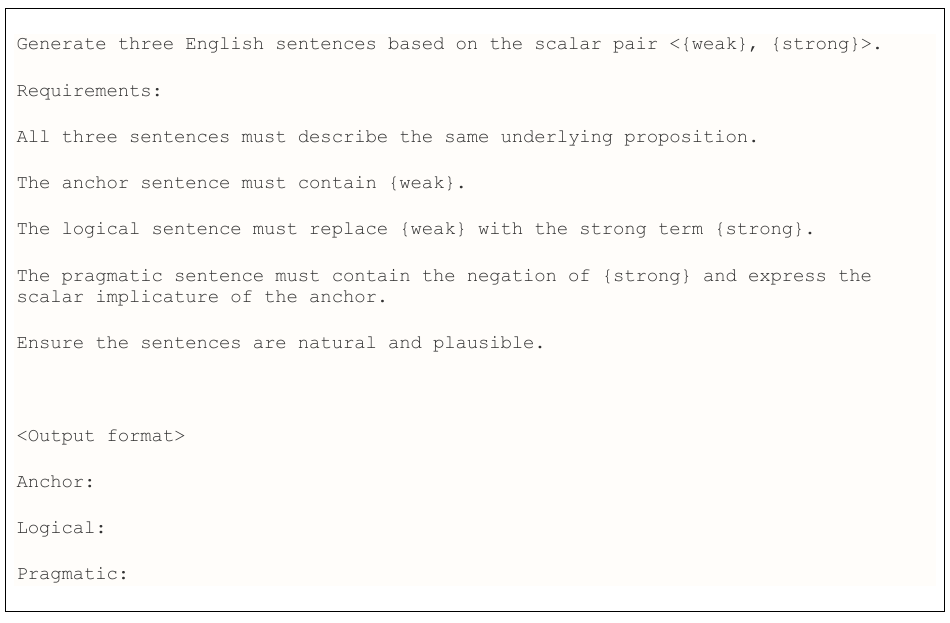}
\end{figure*}

\begin{figure*}
\section{Item-level pragmatic interpretation rates}
\label{app:item_level}
\includegraphics[width=\textwidth, height=0.4\textheight, keepaspectratio]{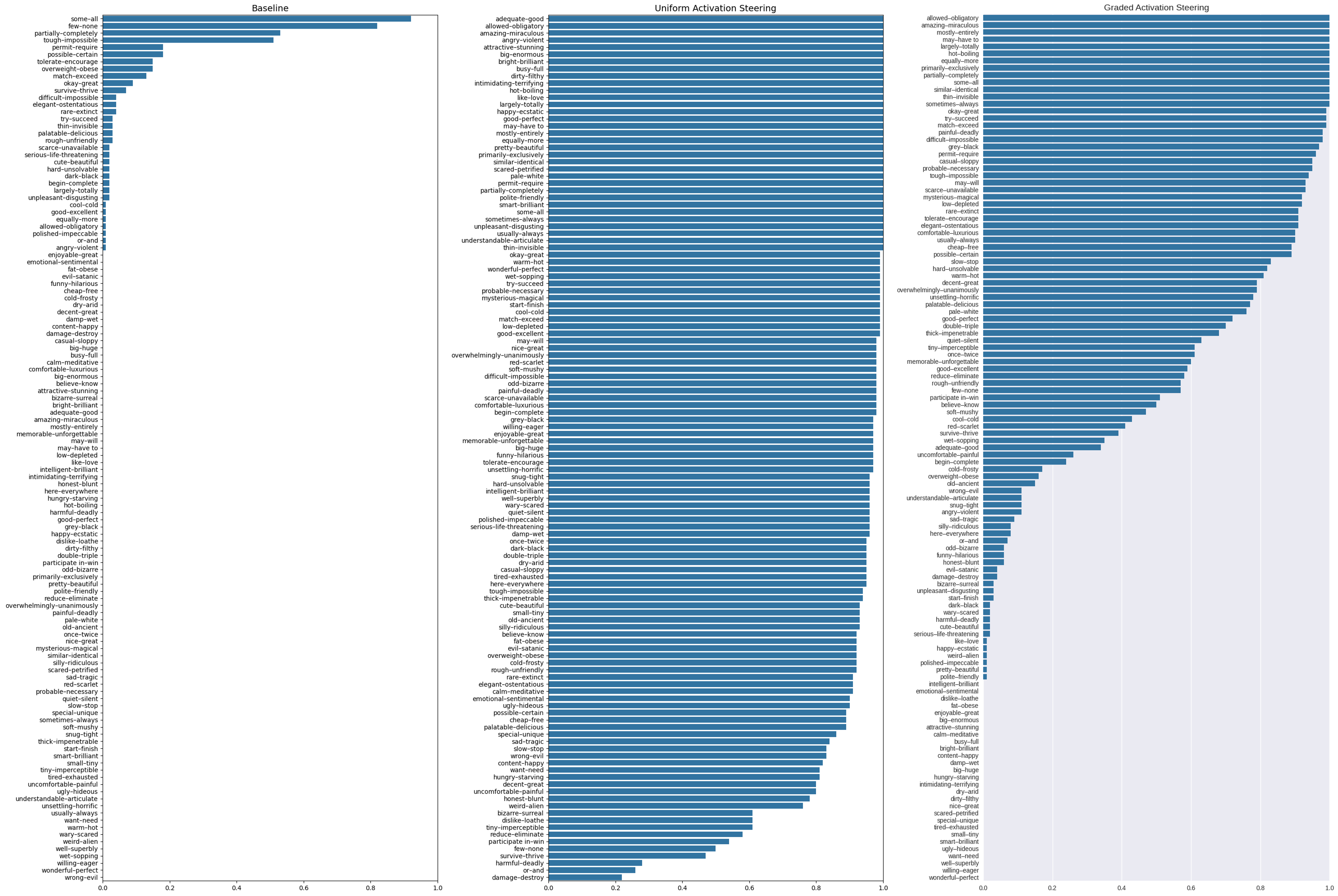}
  \captionsetup{width=\textwidth}
  \caption{Item-level proportions of pragmatic interpretations for LLaMA3 across baseline, uniform activation steering, and graded activation steering conditions}
\end{figure*}

\begin{figure*}
\includegraphics[width=\textwidth, height=0.4\textheight, keepaspectratio]{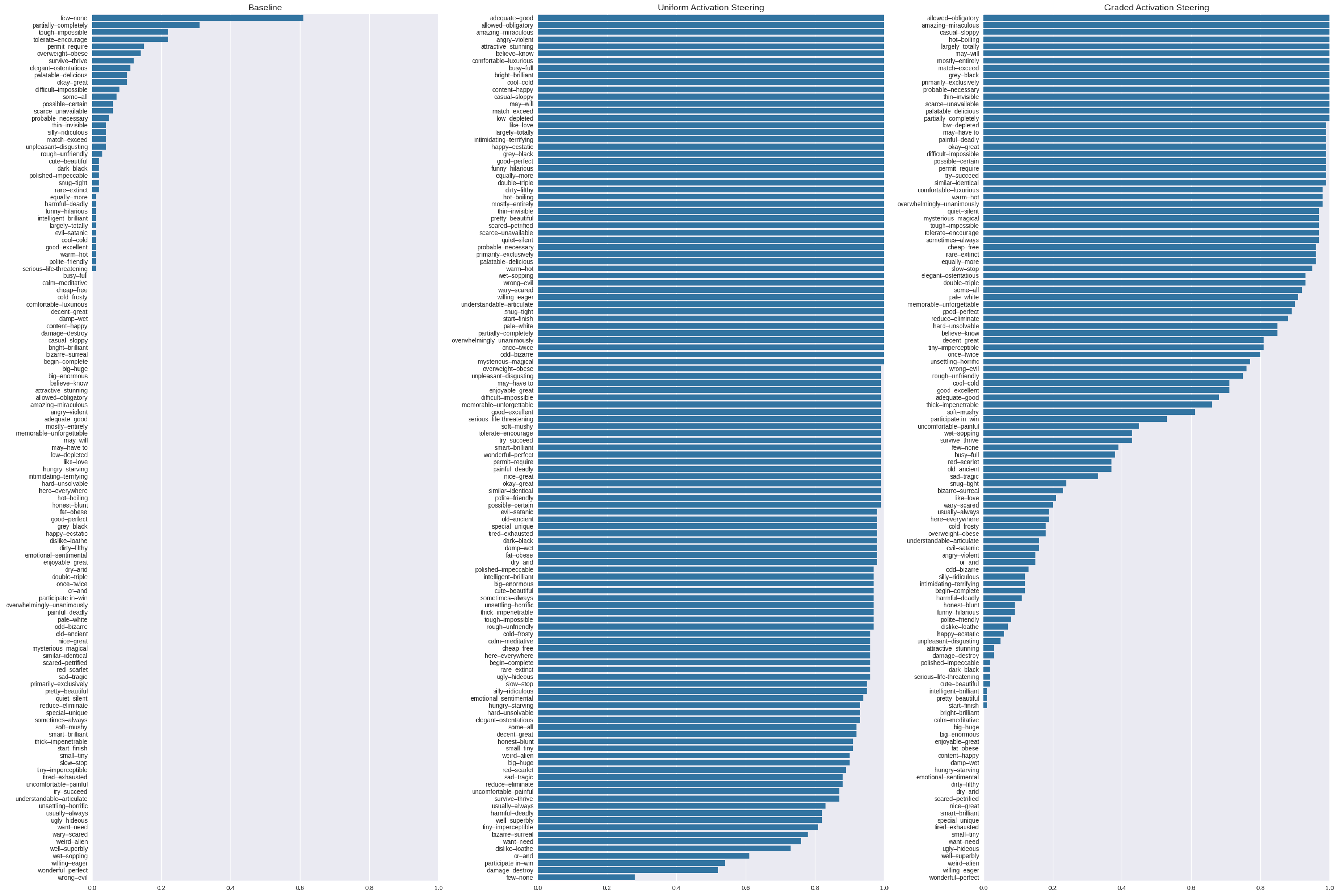}
  \captionsetup{width=\textwidth}
  \caption{Item-level proportions of pragmatic interpretations for Qwen2 across baseline, uniform activation steering, and graded activation steering conditions}
\end{figure*}

\begin{figure*}
  \includegraphics[width=\textwidth]{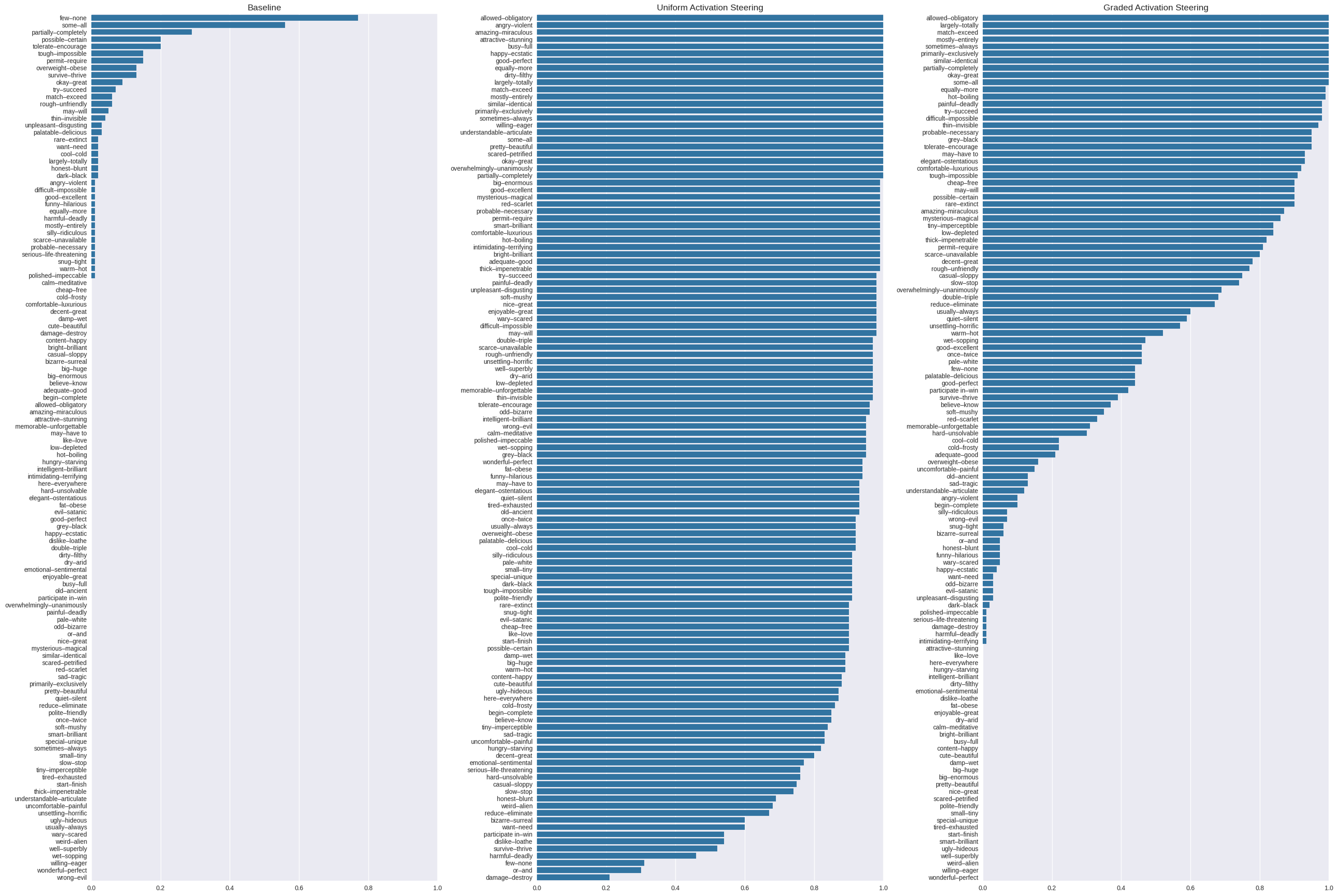}
  \captionsetup{width=\textwidth}
  \caption{Item-level proportions of pragmatic interpretations for Gemma2 across baseline, uniform activation steering, and graded activation steering conditions}
\end{figure*}
\begin{figure*}[p]
  \includegraphics[width=\textwidth]{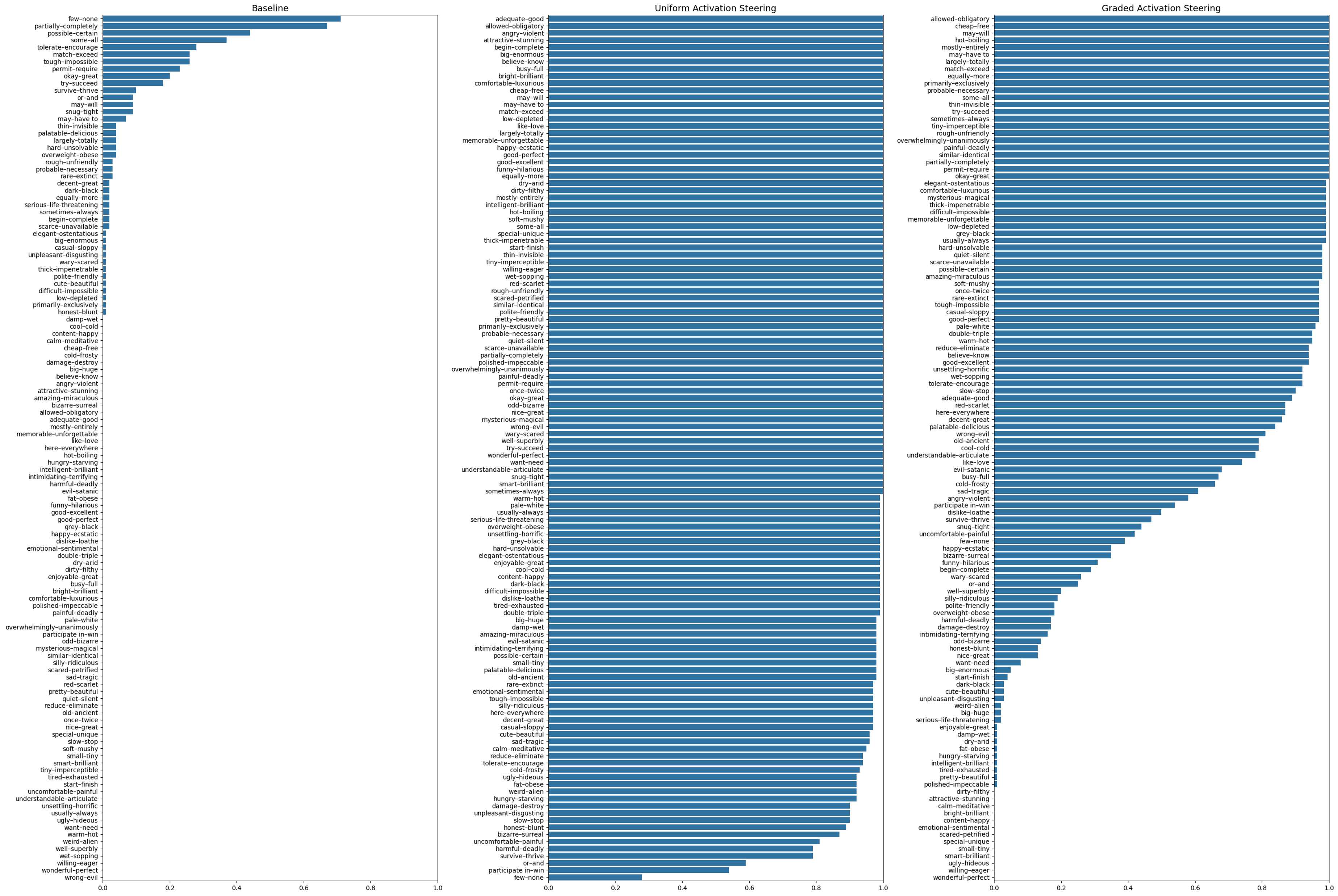}
  \captionsetup{width=\textwidth}
  \caption{Item-level proportions of pragmatic interpretations for OLMo across baseline, uniform activation steering, and graded activation steering conditions}
\end{figure*}

\begin{figure*}
\section{Scatter Plots for Graded Activation Steering}
\label{app:graded}
  \includegraphics[width=\textwidth]{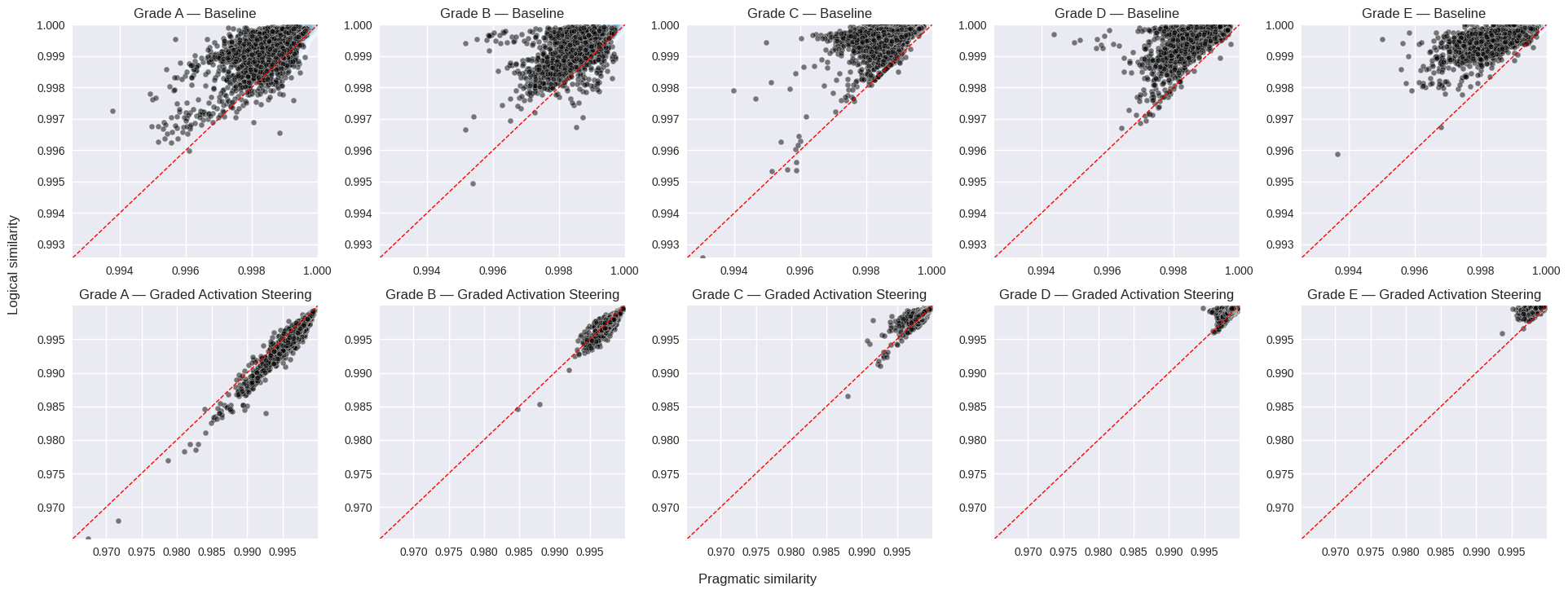}
  \caption{Scatter plots illustrating the relationship between pragmatic similarity (x-axis) and logical similarity (y-axis) under the baseline condition (upper) and graded activation steering (lower) for LLaMA3, based on the graded item set}
\end{figure*}

\begin{figure*}
  \includegraphics[width=\textwidth]{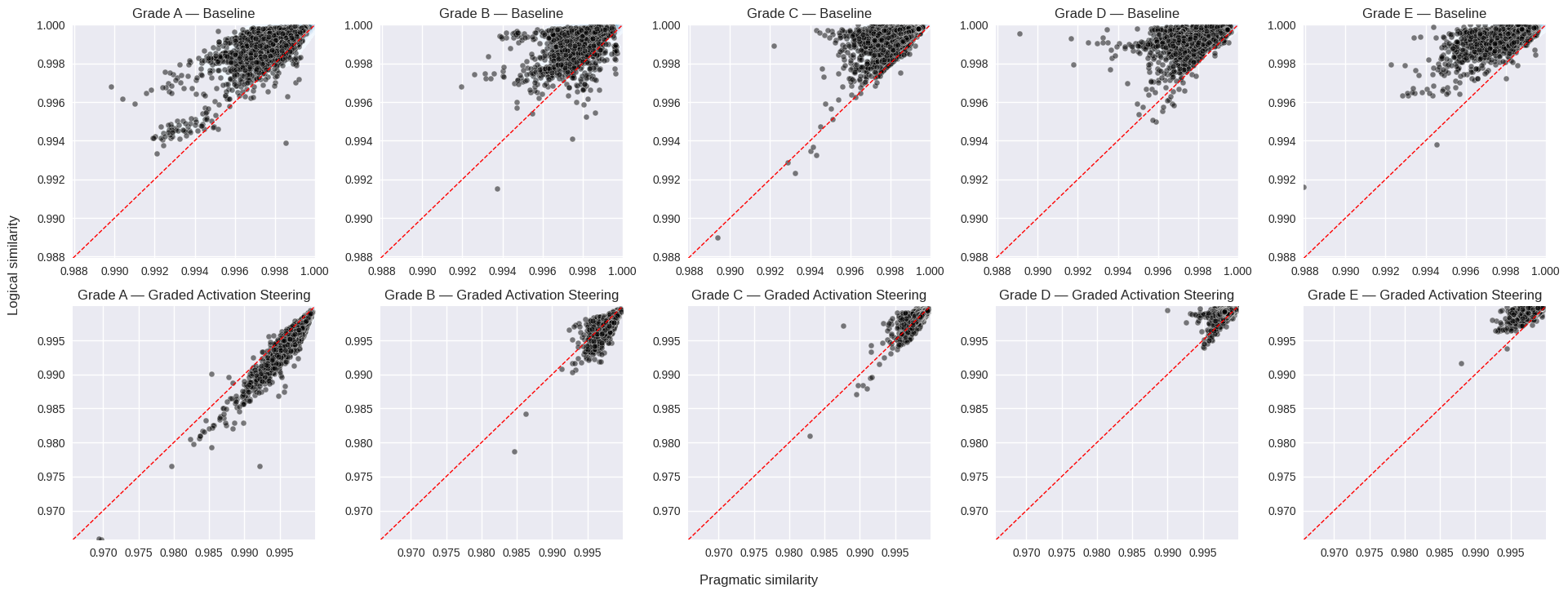}
  \caption{Scatter plots illustrating the relationship between pragmatic similarity (x-axis) and logical similarity (y-axis) under the baseline condition (upper) and graded activation steering (lower) for Qwen2, based on the graded item set}
\end{figure*}

\begin{figure*}
  \includegraphics[width=\textwidth]{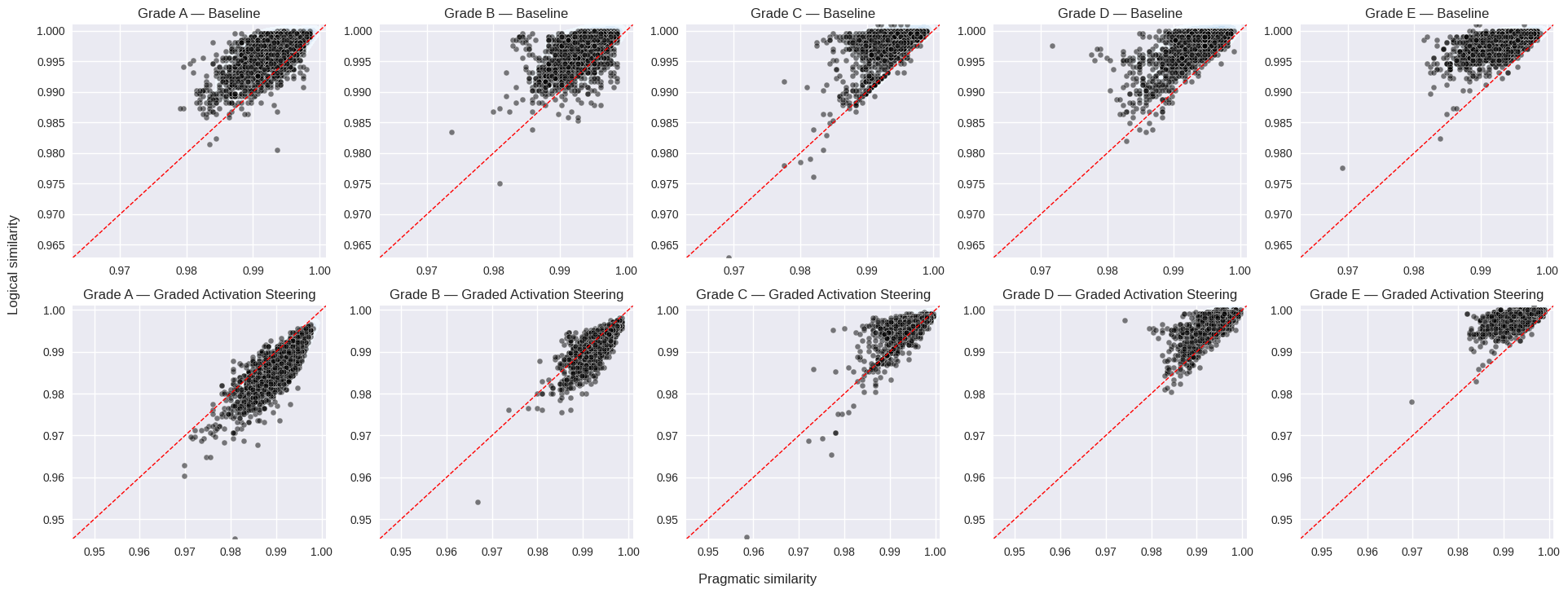}
  \caption{Scatter plots illustrating the relationship between pragmatic similarity (x-axis) and logical similarity (y-axis) under the baseline condition (upper) and graded activation steering (lower) for Gemma2, based on the graded item set}
\end{figure*}

\begin{figure*}
  \includegraphics[width=\textwidth]{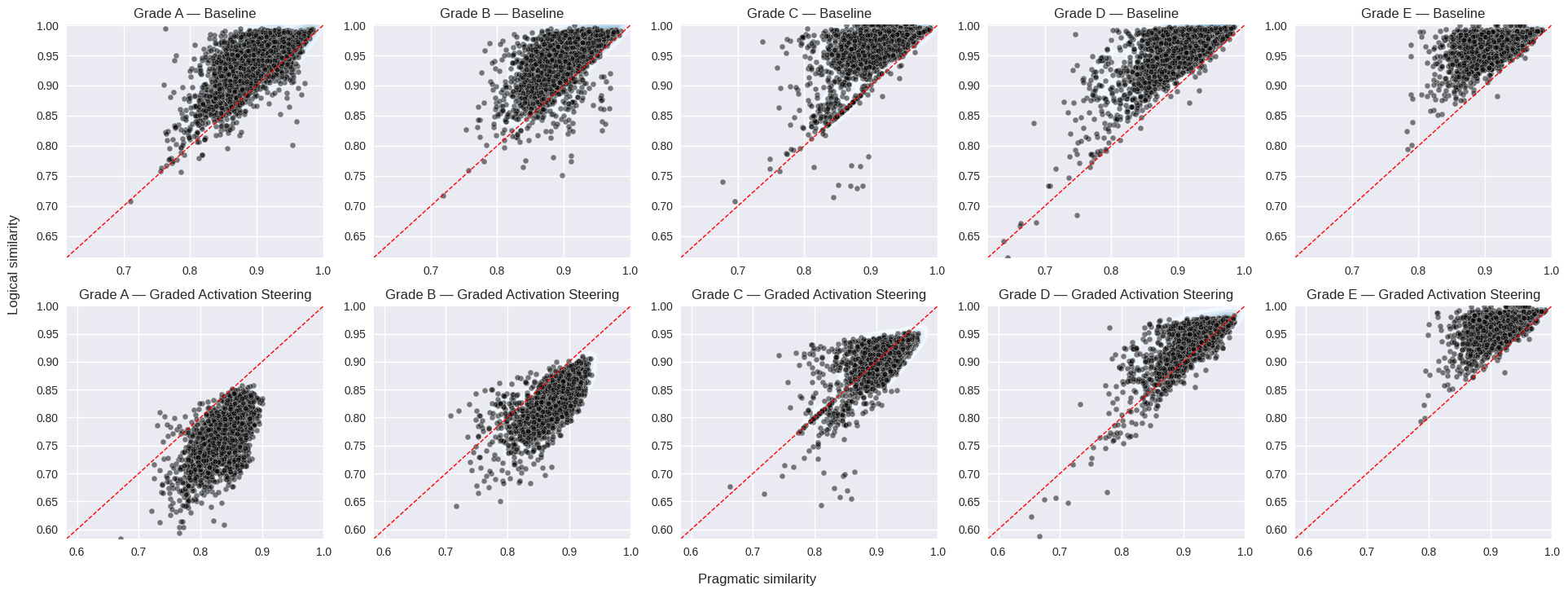}
  \caption{Scatter plots illustrating the relationship between pragmatic similarity (x-axis) and logical similarity (y-axis) under the baseline condition (upper) and graded activation steering (lower) for OLMo, based on the graded item set}
\end{figure*}

\begin{figure*}
\section{Descriptive statistics of item-level deviations under uniform and graded activation steering}
\label{app:deviation}
\vspace{0.5em}
\includegraphics[width=\textwidth]{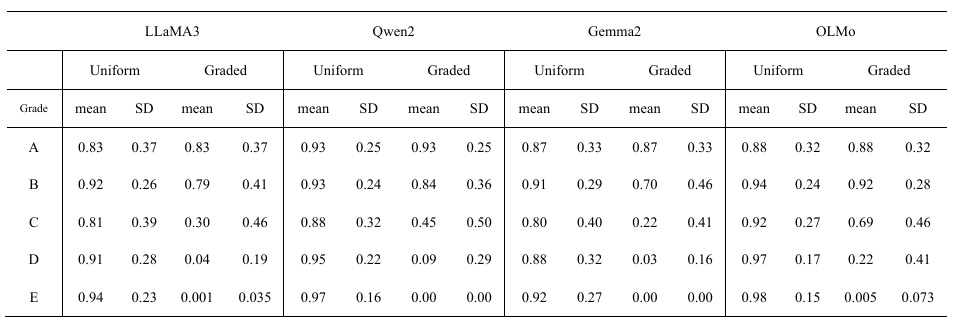}
\end{figure*}

\begin{figure*}[t]
\section{Distributions of item-level changes under uniform and graded activation steering}
\label{app:change}
  \centering
  \begin{minipage}{0.48\textwidth}
    \centering
  \includegraphics[width=\columnwidth]{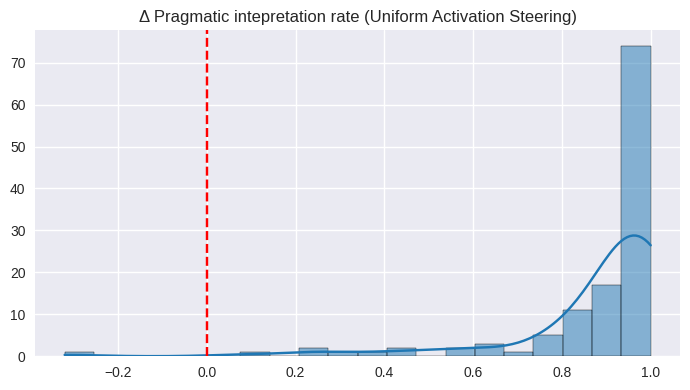}
  \includegraphics[width=\columnwidth]{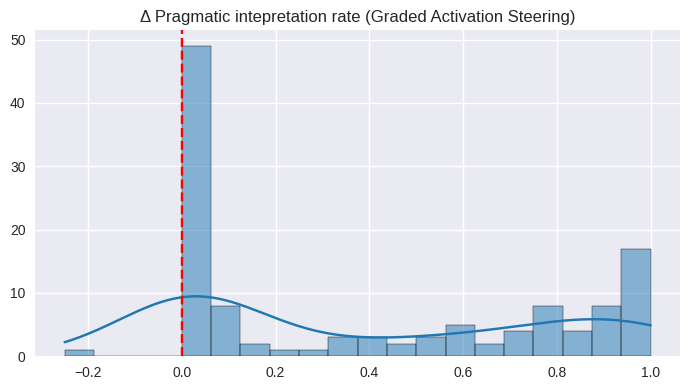}
  \caption{Histograms showing item-level changes in pragmatic interpretation rate ($\Delta$ relative to baseline) under uniform (top) and graded (bottom) activation steering for LLaMA3}
  \end{minipage}
  \hfill
  \begin{minipage}{0.48\textwidth}
  \centering
  \includegraphics[width=\columnwidth]{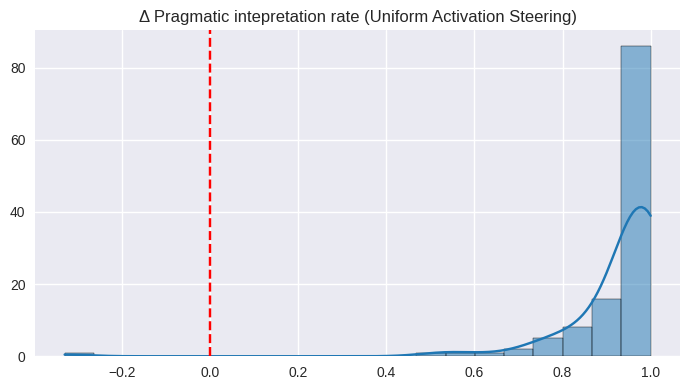}
  \includegraphics[width=\columnwidth]{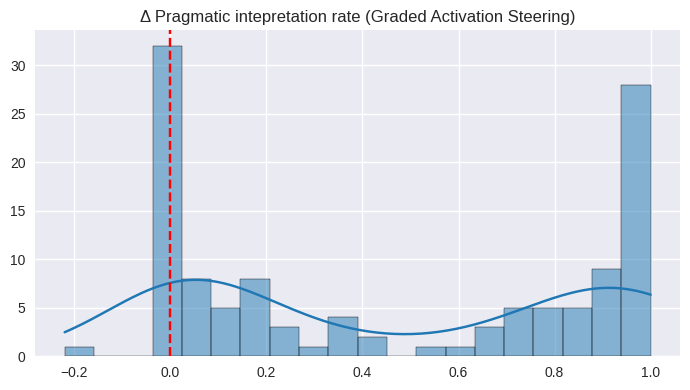}
  \caption{Histograms showing item-level changes in pragmatic interpretation rate ($\Delta$ relative to baseline) under uniform (top) and graded (bottom) activation steering for Qwen2}
  \end{minipage}
\end{figure*}

\begin{figure*}[t]
  \centering
  \begin{minipage}{0.48\textwidth}
    \centering
    \includegraphics[width=\linewidth]{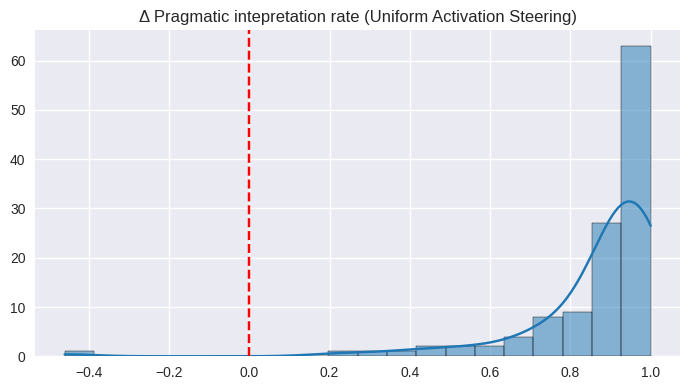}
    \includegraphics[width=\linewidth]{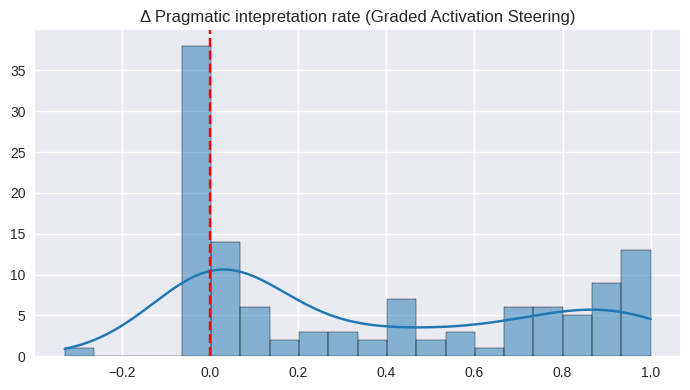}
    \caption{Histograms showing item-level changes in pragmatic interpretation rate ($\Delta$ relative to baseline) under uniform (top) and graded (bottom) activation steering for Gemma2}
  \end{minipage}
  \hfill
  \begin{minipage}{0.48\textwidth}
  \centering
  \includegraphics[width=\columnwidth]{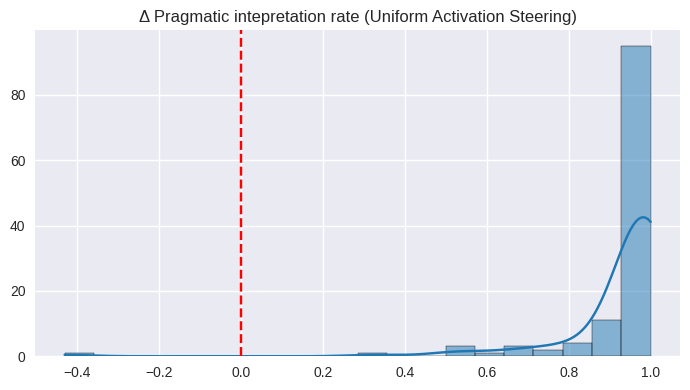}
  \includegraphics[width=\columnwidth]{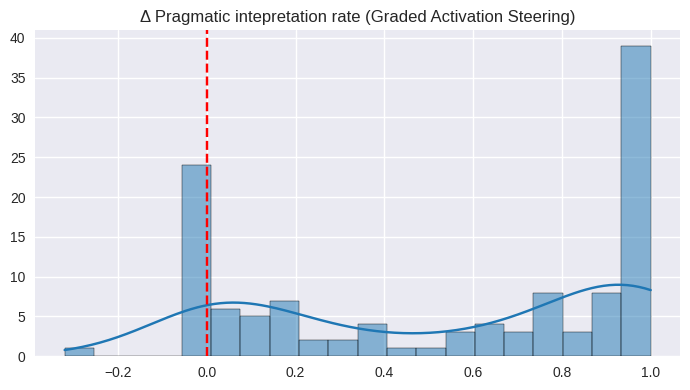}
  \caption{Histograms showing item-level changes in pragmatic interpretation rate ($\Delta$ relative to baseline) under uniform (top) and graded (bottom) activation steering for OLMo}
  \end{minipage}
\end{figure*}

\end{document}